 \documentclass[10pt]{article}
 
 \usepackage{cogsci}

\cogscifinalcopy 

\usepackage{url}
\usepackage{pslatex}
\usepackage{graphicx}
\usepackage[hidelinks]{hyperref}
\usepackage{booktabs}
\usepackage{apacite}
\usepackage{float} 
\usepackage{placeins}
\usepackage{csquotes} 
\usepackage{mdframed} 
\usepackage{xcolor}
\usepackage{todonotes}
\usepackage{siunitx} 
\usepackage{comment}
\usepackage{wrapfig}
\definecolor{Blue}{RGB}{50,50,200}
\definecolor{Red}{RGB}{200,50,50}
\definecolor{Black}{RGB}{0,0,0}

\usepackage{algorithm2e} 
\RestyleAlgo{boxruled}

\newcommand{\numns}[1]{\num[scientific-notation = false]{#1}}

\title{Visual scoping operations for physical assembly}

\author{
\textbf{Felix Binder} \\ Dept. of Cognitive Science \\  UC San Diego \\ \texttt{\href{mailto:fbinder@ucsd.edu}{\url{fbinder@ucsd.edu}}}
 \And
\textbf{Marcelo Mattar} \\ Dept. of Cognitive Science \\  UC San Diego \\ \texttt{\href{mailto:mmattar@ucsd.edu}{\url{mmattar@ucsd.edu}}}
 \And
\textbf{David Kirsh} \\ Dept. of Cognitive Science \\  UC San Diego \\ \texttt{\href{mailto:kirsh@ucsd.edu}{\url{kirsh@ucsd.edu}}}
 \And
\textbf{Judith Fan} \\ Dept. of Psychology \\ UC San Diego \\ \texttt{\href{mailto:jefan@ucsd.edu}{\url{jefan@ucsd.edu}}}
}

\begin{document}

\maketitle

\begin{abstract}
Planning is hard. 
The use of subgoals can make planning more tractable, but selecting these subgoals is computationally costly. 
What algorithms might enable us to reap the benefits of planning using subgoals while minimizing the computational overhead of selecting them? 
We propose visual scoping, a strategy that interleaves planning and acting by alternately defining a spatial region as the next subgoal and selecting actions to achieve it.
We evaluated our visual scoping algorithm on a variety of physical assembly problems against two baselines: planning all subgoals in advance and planning without subgoals.
We found that visual scoping achieves comparable task performance to the subgoal planner while requiring only a fraction of the total computational cost.
Together, these results contribute to our understanding of how humans might make efficient use of cognitive resources to solve complex planning problems.

\end{abstract}

\section{Introduction}

Imagine you are preparing a meal. 
You need to wash, chop, heat, saut\'e, move around, boil. 
How do you know what to do next?
Planning every action in advance is practically impossible: the number of potential sequences of actions grows exponentially with one's action repertoire and the number of steps needed to reach the goal state. 
Not planning at all is no better: some actions must precede others (you can't saut\'e without heating the pan), and timing matters too (e.g., adding sauce to uncooked pasta). 
How do humans manage to routinely solve such complex planning problems in everyday life? 

Classical approaches to planning formulate such problems as search over a space of actions \cite{NewellHumanproblemsolving1972,KirshProblemSolvingSituated2009}, augmented with heuristics and stochastic methods to more selectively search the space of possible actions \cite{GeffnerComputationalmodelsplanning2013}. 
Even with heuristics and stochastic methods, planning complex tasks in a rich environment is computationally prohibitive, because the space of potential plans that needs to be searched grows rapidly with both the number of possible actions available and the number of actions required to achieve the task \cite{Bellman:1957}. 
The computational complexity predicted by the classical approach is at odds with the effortlessness with which people act in the real world \cite{KirshProblemSolvingSituated2009}. 

A promising alternative approach from hierarchical reinforcement learning \cite{botvinick2009hierarchically} permits an agent to learn abstractions over sequences of actions, which can then be invoked as subgoals during planning \cite{MaistoDivideimperasubgoaling2015,ZhangComposablePlanningAttributes2019,BapstObjectorientedstateediting2019}. 
However, while proposing good subgoals can reduce the computational cost of planning the sequences of actions \cite{CorreaResourcerationalTaskDecomposition2020}, actually choosing which subgoals to propose can itself be highly costly. 
How might people manage these costs?

What these formal approaches tend to ignore is that when people confront such tasks they are often embedded in physical environments that can be reconfigured to suit their current goals \cite{Kirshintelligentusespace1995}. 
For example, one could gather all the vegetables on the cutting board and then focus on chopping what is there, ignoring what could be done elsewhere in the kitchen. 
After achieving this subgoal, it may be worth considering what the next subgoal should be (e.g., saut\'eing the vegetables), then focusing on that task (e.g., only considering actions available near the stove), and so on.
We call this \textit{visual scoping}---manipulating the visual environment to select the next subgoal.
While such a strategy would not be expected to always identify the optimal sequence of subgoals, it may reduce the overall cost of jointly inferring subgoals and actions without leading to devastating consequences. 
This paper aims to establish a basic understanding of how such a visual scoping mechanism constrains planning behavior and impacts the overall computational cost of planning. 
Our investigation of visual scoping takes inspiration from recent work exploring how agents intervene on the world to aid in physical reasoning and planning \cite{DasguptaLearningactintegrating2018,Allen29302}. 
Unlike these studies, we focus not on experimental or perceptual interventions which yield new information, but rather on interventions that select already present information to aid physical and spatial reasoning \cite{KirshDistinguishingEpistemicPragmatic1994}. 

As a case study, we consider planning in block-tower reconstruction problems, in which an agent uses an inventory of rectangular blocks to recreate a specific block tower.
The combinatorial nature of construction leads to an explosion in the number of possible states to consider when planning block placements. 
To explore how visual scoping may influence problem solving in this challenging domain, we conducted a set of computational experiments comparing its behavior to that of both classical planning algorithms and more recent approaches which decompose the task into subgoals in advance of action-level planning \cite{CorreaResourcerationalTaskDecomposition2020}.
We discovered that visual scoping can approximate the success and efficiency of full task decomposition on a block-tower reconstruction task while requiring a much smaller computational budget overall. 
We also find that in trading off quick progress and low planning cost in choosing the next subgoals, valuing minimizing planning costs leads to more subgoals and poorer performance.
Together, these findings help advance our knowledge of how perceptual and cognitive constraints interact to support efficient problem solving.

\section{Approach}

\begin{figure}
	\begin{center}
		\includegraphics[width=0.4\textwidth]{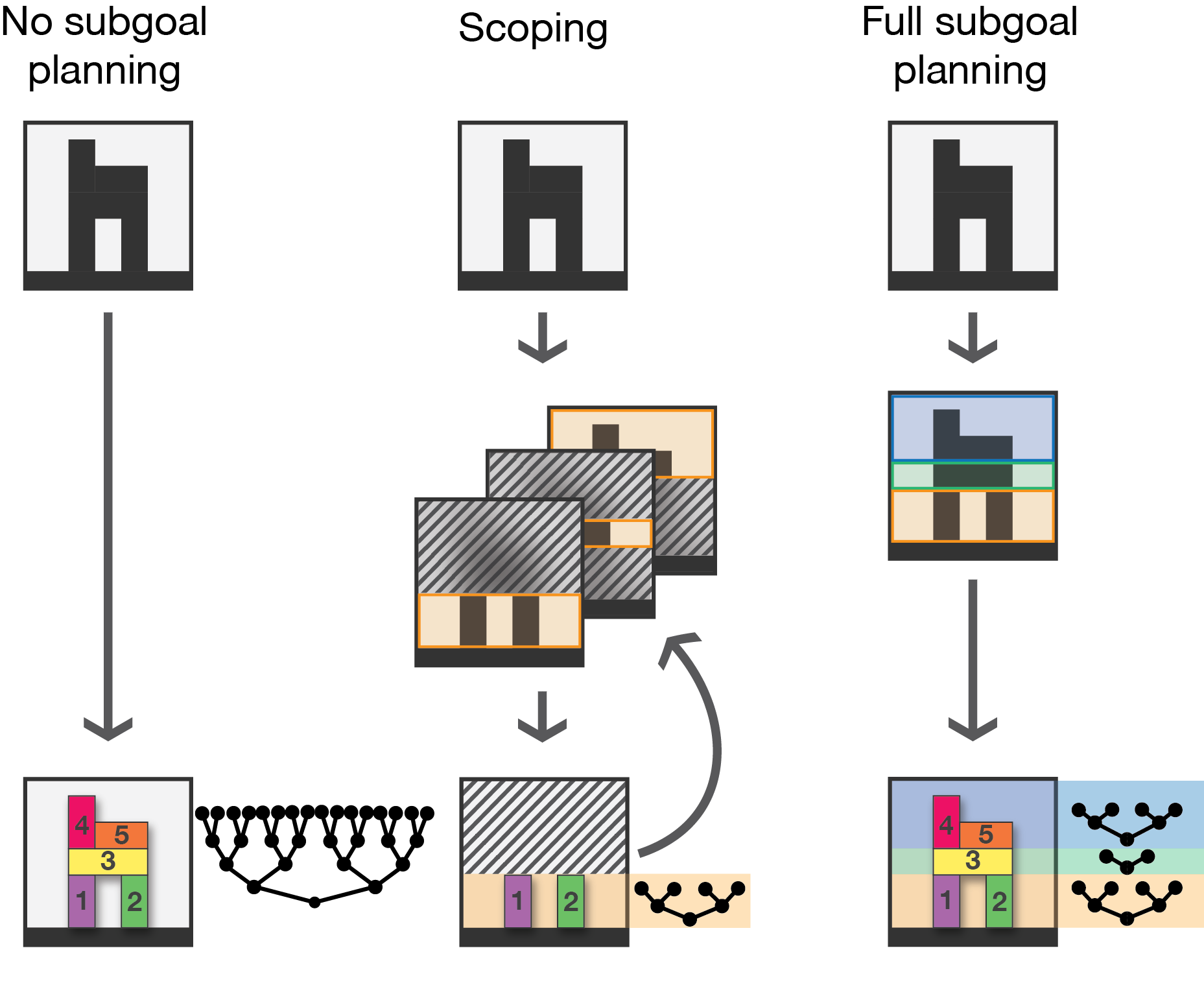}
	\end{center}
	\vspace{-4mm}
	\caption{
		We relate three different strategies for planning using subgoals: no subgoals, planning only the next subgoal (scoping) and planning all subgoals in advance (full-subgoal planning). A schematic search tree of potential actions is shown at the bottom. Both the number of possible different actions (breadth of tree) and the length of sequences of action (depth) increases with the size of the subgoal.
	}
	\label{overview}
\end{figure}


\subsection{Block-tower reconstruction task}

Three specific considerations motivated the choice of block-tower reconstruction for our experiments:
(1) it requires planning over somewhat extended time horizons, (2) it is familiar enough to people that they can easily predict the consequence of their actions, and (3) it is inherently spatial, enabling visual scoping to select regions of the environment to focus on that could constitute plausible subgoals. 
Here we use the \textit{block tower reconstruction task}, which entails assembly of two-dimensional block towers in a gridworld running simulated physics \cite{mccarthylearning}. 
Similar block tower construction tasks have been used to study planning and physical reasoning in artificial agents \cite{Sussmancomputermodelskill1975,BapstStructuredagentsphysical2019} and humans \cite{DietzBuildingblockscomputational2019,CortesaConstraintsDevelopmentChildren2018,mccarthylearning}.

On each trial, the planner is presented with an outline of a shape and has to recreate it by placing blocks from a fixed inventory in a building area. 
An action consists of the choice of a block (eg. 2x1) and a horizontal location. 
The block is then placed on top of the highest block or the ground in that horizontal location. 
Once a block has been placed it cannot be removed, necessitating planning in order to not get stuck in dead ends.\footnote{If actions can be undone, planning is not technically necessary \cite{KirshProblemSolvingSituated2009}: one could just try out actions and backtrack if they fail. However, placing and then undoing physical parts is costly and not always possible. Reasoning about the consequences of actions is thus important, particularly in domains where most paths do not lead to the goal.}
As a simplification in our current experiments, blocks were also \enquote{glued down} after they were placed, preventing towers from toppling over. 
The trial ends either when the target shape is perfectly reconstructed, when no further block can be placed or when the planner chooses to not place a further block. The trial is considered successful if the target shape has been exactly reconstructed.
16 unique silhouettes were used in our experiments%
, spanning a range of difficulty levels for the planner. 
This is a challenging task for human participants: human subjects average 22.4\% perfect reconstruction, albeit without glued blocks \cite{mccarthylearning}. 

\subsection{Visual scoping}

The use of subgoals can reduce both the number of possible different actions and the length of sequences of actions needed to reach the (sub)goal (see Figure \ref{overview}). Visual scoping attempts to reap this computational benefit while minimizing the cost of subgoal selection by only choosing the next subgoal depending on the current state of the environment.
Our visual scoping planner works by interleaving two operations: (1) identifying the next subgoal to achieve, and (2) using an action-level search algorithm to plan the sequence of actions to achieve that immediate subgoal (Algorithm~\ref{combo_alg}, Figure~\ref{overview}).
Subgoals are defined spatially: a subgoal is defined as a region of space rather than as a specific state of the world. 
The set of potential subgoals we consider are horizontal layers, eg. the first or the first three rows of the target shape. 
Splitting a construction problem into horizontal layers is sensible given the incremental nature of building: it is easier to place blocks on top of other blocks rather than below them. 
\citeA{CortesaCharacterizingspatialconstruction2017} show that young children naturally adopt a layerwise strategy on a related block construction task.
In our block-tower reconstruction setting, this corresponds to identifying the height of the subtower the agent seeks to reconstruct, then searching over potential action plans for doing so.

\subsection{Planners}

\begin{algorithm}
	\caption{Full-subgoal \& scoping planner}
	\label{combo_alg}
	\small
		Input: action-level search algorithm $LLP$, target shape $G$, environment $E$\\ 
		Parameters: weight parameter $\lambda$, computational budget $b$\\
	$c^{total} \leftarrow 0$\\
	~\\
	\begin{mdframed}[rightmargin = 20pt,skipbelow = 0pt]
		Run once:   \null\hfill \textit{Full}
	\end{mdframed} 
	\vspace{-4pt}
	\begin{mdframed}[rightmargin = 20pt, skipabove = 0pt]
		While $ E \neq G$:   \null\hfill \textit{Scoping}
	\end{mdframed}
	\Indp
		\begin{mdframed}[leftmargin=\tabcolsep, rightmargin = 20pt,skipbelow = 0pt]
			$\Phi \leftarrow \{(g_{1:m1}), (g_{1:m2}),\dots,(g_{1:mn})~where~g_{m1}=g_{m2} = \dots =g_{nm} = G\}$ \textcolor{gray}{possible complete sequences of subgoals given $E$}   \null\hfill \textit{Full}
		\end{mdframed} 
		\vspace{-4pt}
		\begin{mdframed}[leftmargin=\tabcolsep, rightmargin = 20pt, skipabove = 0pt]
			$\Phi \leftarrow \{(g_1), (g_2),\dots,(g_n)~where~g_1,g_2,\dots,g_n > E\}$ \textcolor{gray}{possible sequences of subgoals given $E$ of length $1$} \\ \null\hfill \textit{Scoping} 
		\end{mdframed}		
	For each $(g)$ in $\Phi$:\\
	\Indp
		$s$ = Empty\\ 
		For each $g$ in $(g)$:\\
		\Indp
			$r_g$ $\leftarrow$ $Area(g)-Area(s)$ \null\hfill \\
			While $s_{final} \neq g$  and $c^{planning}_g < b$:\\ 
			\Indp
					$c^{solution} = 0$\\ 
				  $[(a_1,\dots,a_l)_g,~c^{solution}_g,~s_{final}] \leftarrow LLP(s,g)$\\
				  $c^{planning}_g~+=~c^{solution}_g$\\
		  \Indm
		  $c^{total} += c^{planning}_g$\\
		  if $s_{final} = g$: \textcolor{gray}{if subgoal is solved} \\ 
		  \Indp
			  $s$ $\leftarrow$ $s_{final}$ \null\hfill\\  
	  \Indm
		  else:\\
		  \Indp
			   $c^{planning}_g = \infty$\\
			   Continue to next $(g_k) \in \Phi$\\
		  \Indm \Indm \Indm
		  For each $(g)$ in $\Phi$:\\
		  \Indp
				  $V_{(g)} \leftarrow$ $\sum_{g \in (g)} r_g - \lambda *c^{planning}_g $\\
		  \Indm
		  \begin{mdframed}[leftmargin=\tabcolsep, rightmargin = 20pt,skipbelow = 0pt]
			Apply $(a_1,\dots,a_l)_{g \in argmax(V_{(g)})}$ to $E$ \textcolor{gray}{Choose sequence with highest score and apply all actions} \hfill\textit{Full}
		\end{mdframed} 
		\vspace{-4pt}
		\begin{mdframed}[leftmargin=\tabcolsep, rightmargin = 20pt, skipabove = 0pt]
			Apply $(a_1)_{g \in argmax(V_{(g)})}$ to $E$ \textcolor{gray}{Choose sequence with highest score and apply only the first action}  \hfill \textit{Scoping}
		\end{mdframed}
		\Indm
	  define $LLP(s,g)$:\\
	  \Indp
		  \textcolor{gray}{Attempts to find a sequence of actions from $s$ to $g$}\\
		  return $[(a_1,a_2,\dots,a_n), c^{solution}, s^{final}]$\
  \end{algorithm}

\subsubsection{Subgoal level planners}
We implemented two hierarchical subgoal planners, scoping and full-subgoal planning (see Algorithm \ref{combo_alg} for a comparison of the algorithms), as well as a baseline of not using subgoals at all. 
The \textbf{scoping planner} considers all potential next subgoals given the current state of the environment. After a subgoal has been chosen, the subgoal is passed on to the action-level search algorithms and the actions it returns are applied to the environment. The process repeats until the target shape is completed or no subgoal that is perfectly solvable can be found.  
It has to trade off immediate progress and computational cost when selecting a subgoal: does it prefer more substantial subgoals (i.e., making rapid progress) even at the cost of higher computational cost, or more modest subgoals that are easier to solve?
This tradeoff is controlled by the $\lambda$ parameter. 
The higher the value of $\lambda$, the more the scoping planner works to minimize action-planning cost. 
When $\lambda = 0$, the scoping planner maximizes progress no matter the cost.
%
The \textbf{full-subgoal planner} finds one sequence of subgoals from the beginning of building to the final structure in one go. It first considers all possible sequences of subgoals that end in the full target shape, selects the one that minimizes the action-level search algorithm computational cost and builds the structure from start to finish, as in \citeA{CorreaResourcerationalTaskDecomposition2020}. 
%
Finally, to compare the use of subgoal planners to a baseline of \textbf{pure action-level search} not using subgoals at all we also apply the pure action-level search algorithm directly on the target structure without decomposing it into subgoals. 

These planners use a given action-level search algorithm to determine for each potential subgoal the computational cost of solving the subgoal with the action-level search algorithm. 
This is done by running the action-level search algorithm repeatedly on a subgoal until a solution is found or the cost threshold $b$ is exceeded. It is necessary to sample repeatedly, since the action-level search algorithms used here break ties randomly between equally good plans and therefore might yield different results. 
Since not all subgoals are achievable for a certain action-level search algorithm, a threshold is needed after which the subgoal is considered unsolvable.
Only solvable subgoals are considered by the subgoal level planner. 
Finally, the subgoal planner selects the subgoal or sequence of subgoals that maximizes progress (the proportion of the target shape solved) while minimizing the cost of solving the subgoals using the action-level search algorithm. 
Because the subgoal planners minimize the planning cost of the action-level search algorithm, they perform resource-rational task decomposition. \cite{CorreaResourcerationalTaskDecomposition2020}.
Our implementation of resource-rational task decomposition differs from \citeA{CorreaResourcerationalTaskDecomposition2020} in three specific ways: (1) the use of abstract subgoal states (rather than specific world states serving as subgoals), (2) the ability to only plan a certain number of steps ahead as mediated through $\lambda$, and (3) the ability to handle impossible subgoals and stochastic action-level search algorithms.

\subsubsection{Action-level search algorithms}
The subgoals planners are hierarchical agents: they find a subgoal decomposition both using a particular action-level search algorithm, and specifically for that particular algorithm.
An action-level search algorithm searches the space of possible actions to find a sequence of actions to reach a certain goal. 
Since the space of potential states (different placements of blocks in the building area) is very large---there are roughly  $35^{15} \approx 1.5\times 10^{23}$ different states---exhaustively searching the entire space of states for a complete sequence of actions is not feasible. 
Therefore, the action-level search algorithms are implemented as lookahead planners: if a path to the solution cannot be found within a certain computational budget, the planner plans a sequence of actions as long as it can given a computational constraint with ties broken randomly, then takes the first action of that sequence and plans again, now on the basis of the state resulting from the action just taken. 
We chose two classical search algorithms: one performing brute search, the other performing search informed by a heuristic.

\textbf{Breadth first search lookahead} (BFS) exhaustively explores all possible placements of blocks in the target shape $n$ steps into the future or until a perfect reconstruction is found, then chooses the sequence of actions that maximizes the area of the target shape filled out. 

\textbf{A* lookahead} runs the A* search algorithm \cite{HartFormalBasisHeuristic1968} either until a sequence of actions resulting in a perfect reconstruction is found or until a specified computational budget is exceeded.
A* chooses to explore the action that minimizes $f(s) = g(s) + h(s)$, where $g(s)$ is the cost of reaching the current state from the start measured in number of blocks placed down and $h(s)$ is the heuristic for how expensive the target is to reach from the current state. Here, $h(n)$ is the number of cells left in the target shape divided by the size of the smallest block, so it provides a upper bound for the number of blocks left to fill out the target shape.\footnote{Note that the heuristic $h(s)$ is not strictly admissible: for states that cannot possible reach a perfect reconstruction (for example, if a hole has been left and covered by a block), this locked in dead end is not going to be apparent in the heuristic.}  

\subsubsection{Hypotheses}
The full-subgoal planners considers the largest set of subgoal decompositions, so we expect that it has the highest rate of perfect reconstruction as well as the lowest action planning cost of the solutions themselves compared to the scoping planner. 
However, we hypothesize that this optimality comes at a steep cost: we expect the subgoal-planning cost incurred by the full-subgoal planner to be higher than that incurred by the other two.
By contrast, we predict that the scoping planner will trade off higher performance and cheaper action-level planning for a large decrease in the cost of finding the subgoals. 
Finally, we predict that not using subgoals at all will lead to a much lower rate of perfect reconstruction and higher action-level algorithmic cost of the solution that is found. 

\subsubsection{Experiments}
We ran each planner $32$ times on each of the $16$ structures using different random seeds. For the scoping planner, we used $64$ different values of $\lambda$ each.

\section{Results} 

\begin{figure}[h]
	\begin{center}
		\includegraphics[width=0.45\textwidth]{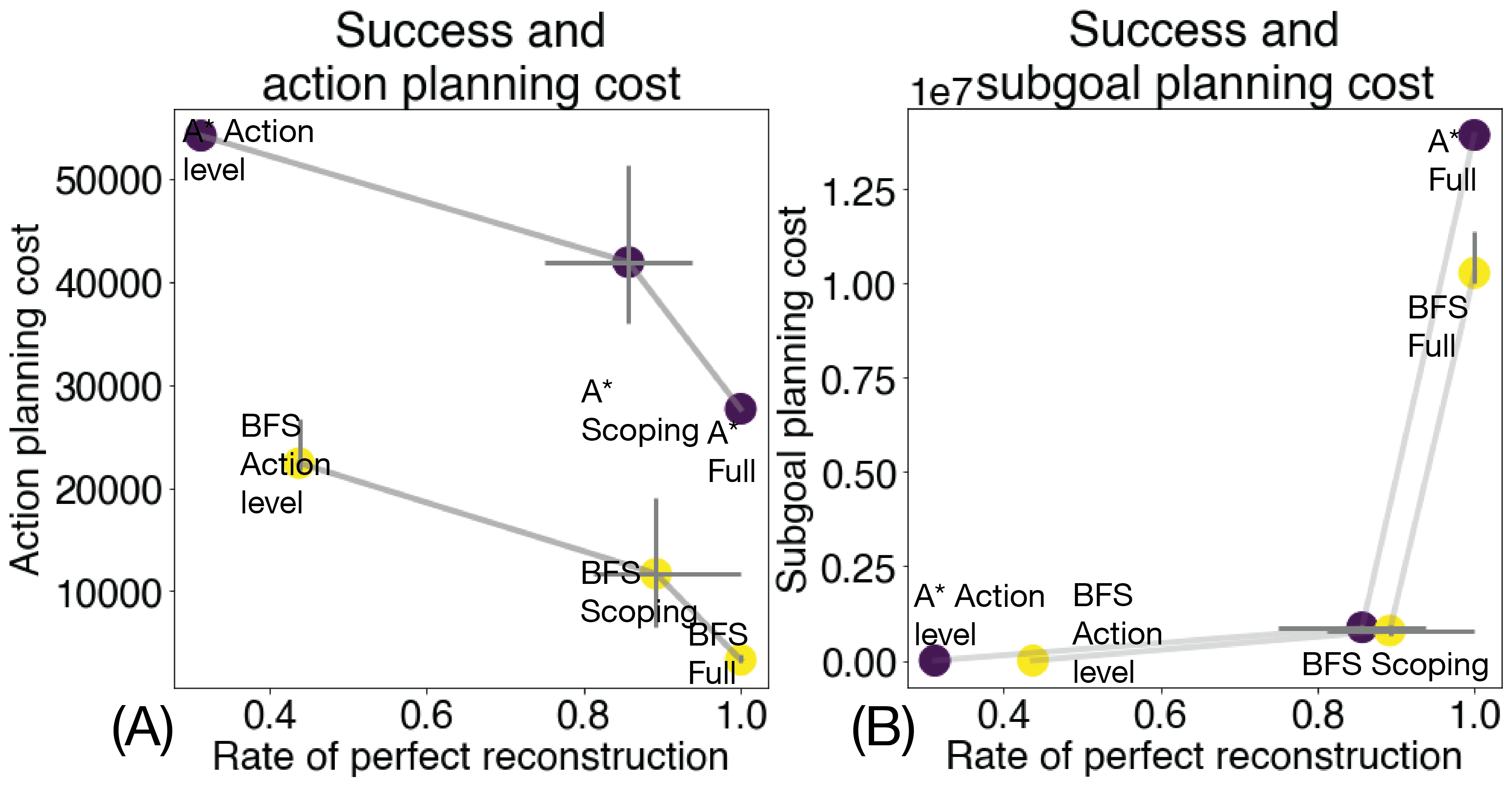}
	\end{center}
	\vspace{-5mm}
	\caption{
		Scoping can approximate the success and action planning cost of full-subgoal planning while requiring a much smaller cost in choosing subgoals.
		(A) Each dot corresponds to a combination of subgoal planner (no subgoals, scoping, full-subgoal planning) and action-level search algorithm (A*, BFS). The cost of planning the actions given the sequence of subgoals and the rate of perfect reconstruction is shown. 
		(B) Same as (A), but with the cost of selecting the subgoals shown instead.
		Costs are measured in evaluated number of states.
		Error bars represent bootstrapped 95\% confidence intervals.
		 } 
	\label{scatter_success_cost}
\end{figure}

\subsubsection{Pure action-level search}

To establish a baseline for both success and planning cost, the action-level search algorithms were used to try to find solutions to the 16 target shapes. Table \ref{comparision_action_planners} shows the success and planning cost of solutions found for a number of configurations of BFS lookahead and A* lookahead without resampling. 
Two particular action-level search algorithms will be used in conjunction with the scoping and full-subgoal planner: BFS lookahead with a search depth of $3$ and A* lookahead with a budget of $4096$ iterations. Using pure-action level search with resampling, BFS achieves a rate of perfect reconstruction of 
$\numns{0.4375}$ ($95\%$ confidence interval (CI): $[\numns{0.4375}, \numns{0.4375}]$; CI derived by computing the mean reconstruction rate across towers on each of 1000 iterations, where each iteration is defined by a novel permutation of the random seeds used to initialize search); 
A* achieves $\numns{0.3125}$ ($95\%$ CI: $[\numns{0.3125},\numns{0.3125}]$).
Block construction is a challenging task: pure action-level search is far from performing at ceiling.
Increasing the search budget of the action-level search algorithms leads to an increase in success, but with it comes a ballooning of planning costs.

\begin{table}
	\centering 
  \resizebox{0.47\textwidth}{!}{
	\begin{tabular}{@{}llllll@{}}
		\toprule
		\textbf{Search} & \textbf{Budget} & \textbf{Accuracy} & \textbf{95\% CI} & \textbf{Cost} & \textbf{95\% CI} \\ \midrule
		Random  & None & \numns{0.0383125} & [\numns{0.0}, \numns{0.125}] & 0 & [0, 0] \\ \midrule
		BFS  & Depth 1 & \numns{0.099625} & [\numns{0.0}, \numns{0.25}] & \num{82.9436} & [\num{82.0}, \num{84.8}] \\ \midrule
		BFS  & Depth 2 & \numns{0.186562} & [\numns{0.0625}, \numns{0.3125}] & \num{1520.14} & [\num{1447.0}, \num{1593.25}] \\ \midrule
		BFS  & Depth 3 & \numns{0.263188} & [\numns{0.125}, \numns{0.4375}] & \num{24544.5} & [\num{22164.994444444445}, \num{26918.513888888887}] \\ \midrule
		BFS  & Depth 4 & \numns{0.41} & [\numns{0.1875}, \numns{0.625}] & \num{351777} & [\num{314981.30333333334}, \num{383280.9683333333}] \\ \midrule
		A* & $8$ Iterations & \numns{0.0615} & [\numns{0.0}, \numns{0.1875}] & \num{120.15} & [\num{114.75}, \num{126.0}] \\ \midrule
		A* & $64$ Iterations & \numns{0.083375} & [\numns{0.0}, \numns{0.1875}] & \num{342.018} & [\num{327.6666666666667}, \num{370.6666666666667}] \\ \midrule
		A* & $4096$ Iterations & \numns{0.122563} & [\numns{0.0}, \numns{0.25}] & \num{16533.2} & [\num{14530.111111111111}, \num{18527.88888888889}] \\ \midrule
		A* & $65536$ Iterations & \numns{0.100125} & [\numns{0.0}, \numns{0.25}] & \num{157444} & [\num{139316.33333333334}, \num{173152.33333333334}] \\ \bottomrule
	\end{tabular}}
  \caption{Increasing the budget of action-level search increases success somewhat, but costs grow exponentially. The table shows the cost and accuracy of various action-level search algorithms without resampling on a single attempt on the block construction task.}  
  \label{comparision_action_planners}
\end{table}

\subsubsection{Full-subgoal planning}
We sought to replicate the finding that using subgoals in planning can reduce action planning cost \cite{CorreaResourcerationalTaskDecomposition2020}. 
We apply the full-subgoal planner to the same $16$ structures, using both BFS lookahead and A* lookahead as action-level search algorithms. 
The full-subgoal planner achieves a perfect reconstruction on every attempt. The action planning cost of those solutions is cheaper compared to pure action-level search 
(\num{19030.836151785716} fewer states evaluated, $95\%$ CI: $[\num{17898.98191964},\num{23536.66607143}]$, $p<0.001$;
A*: \num{26506.25}, $95\%$ CI: $[\num{26506.25},\num{26506.25}]$, $p<0.001$; CI derived by computing the mean paired difference in planning cost between agents on each of 1000 iterations, where each iteration is defined by a new set of random seeds) 
 compared to the respective action-level search algorithm without subgoals. 
While the chosen subgoals themselves are easy to solve, the cost of coming up with the sequence of subgoals is very large (BFS: $M= \num{1.02807e+07}$ states evaluated, $95\%$ CI: $[\num{10007232.0},\num{11394938.4109375}]$;
 A*: $\num{1.39378e+07}$, $95\%$ CI: $[\num{13937763.0625},\num{13937763.0625}]$).

\subsubsection{Scoping}

\begin{figure}
	\begin{center}
		\includegraphics[width=0.5\textwidth]{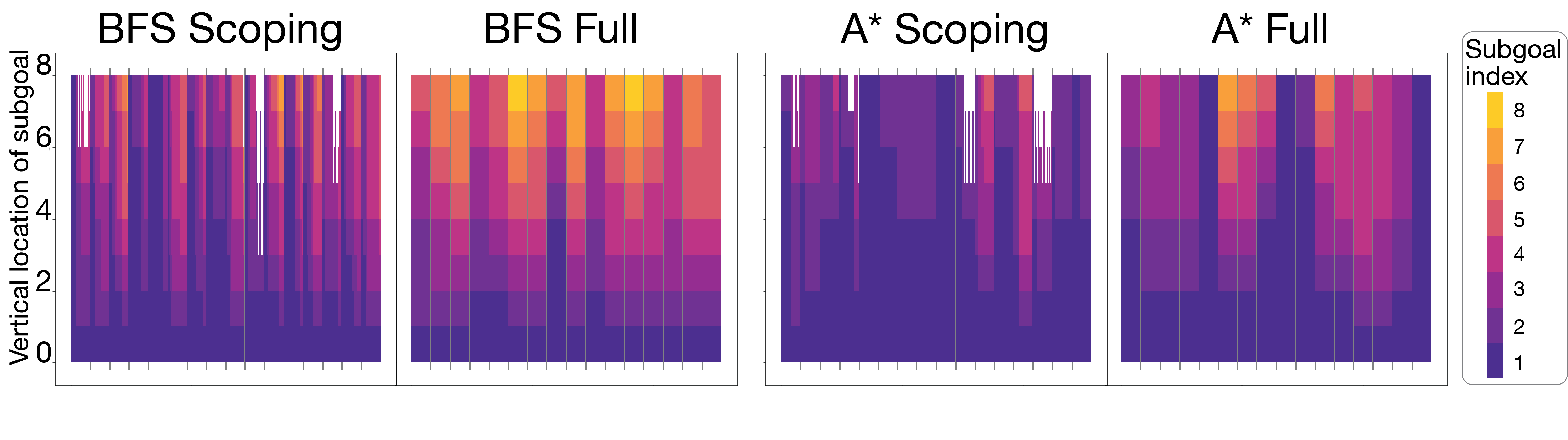}
	\end{center}
	\vspace{-5mm}
	\caption{An overview over the different decompositions chosen by each planner. Each vertical stripe corresponds to a single attempt of the planner to build a certain target shape. Each chosen subgoal is marked in a different color, with blue indicating a the first and orange the eight chosen subgoal. The columns are organized by target shape and for the scoping planners sorted by $\lambda$.} 
	\label{sandplot}
\end{figure}

To capture the planning behavior over a broad range of values for $\lambda$, we average over the dynamic range of  $\lambda$ (BFS: $[0,0.008]$, A*: $[0,0.003]$) for the following analysis.
The scoping planner achieves a slightly lower success compared to full-subgoal decompositions perfect success rate 
(BFS: $\numns{0.104}$ lower rate of perfect reconstruction compared to full-subgoal planner, $95\%$ CI: $[\numns{0.},\numns{0.1875}]$, $p=\numns{0.0415}$; 
A*: $\numns{0.145375}$, $95\%$ CI: $[\numns{0.0625},\numns{0.25}]$, $p=\numns{0.0075}$).
The scoping planner finds more expensive solutions
(BFS: \num{9107.2985625} more states evaluated, $95\%$ CI: $[\num{3767.0765625},\num{17013.9140625}]$, $p<0.001$; 
A*: \num{14842.3570625}, $95\%$ CI: $[\num{8268.4375},\num{25458.18125}]$, $p<0.001$). 
However, the subgoal planning cost is dramatically lower  
(BFS: \num{9527501.5608125} fewer states evaluated, $95\%$ CI: $[\num{9079226.628125},\num{10673451.0546875}]$, $p<0.001$;
A*: \num{13060800.5558125}, $95\%$ CI: $[\num{12992823.25},\num{13161382.36875}]$, $p<0.001$)
---see Figure \ref{scatter_success_cost}.
Figure \ref{sandplot} shows the chosen task decompositions made by the planners.

When building block towers, the planner can choose to use large blocks, making progress quickly, or to apply a more conservative strategy of using smaller blocks. The average number of blocks in a successful solution measures this tendency.
Using BFS lookahead, the scoping planner uses more blocks on average compared to pure action-level search (\num{5.3305625} more blocks used per solution, $95\%$ CI: $[\num{2.6875},\num{7.875}]$, $p<0.001$), but insignificantly fewer than the full-subgoal planner (\num{2.2724375} fewer blocks, $95\%$ CI: $[\num{-1.6890625},\num{6.0640625}]$, $p=\numns{0.064}$). Thus, the scoping planner using BFS lookahead finds somewhat longer solution solution compared to pure action-level search, but not full subgoal decomposition.
Under A* lookahead, the scoping planner uses fewer blocks than both pure action-level search (\num{1.4135} fewer blocks, $95\%$ CI: $[\num{-0.25},\num{3.3125}]$, $p=\numns{0.027}$) and full-subgoal planning (\num{3.0954375} fewer blocks, $95\%$ CI: $[\num{1.5625e-03},\num{6.1875e+00}]$, $p=\numns{0.015}$).

\subsubsection{Trading off immediate progress and action planning cost}

\begin{figure*}
	\begin{center}
		\includegraphics[width=0.94\textwidth]{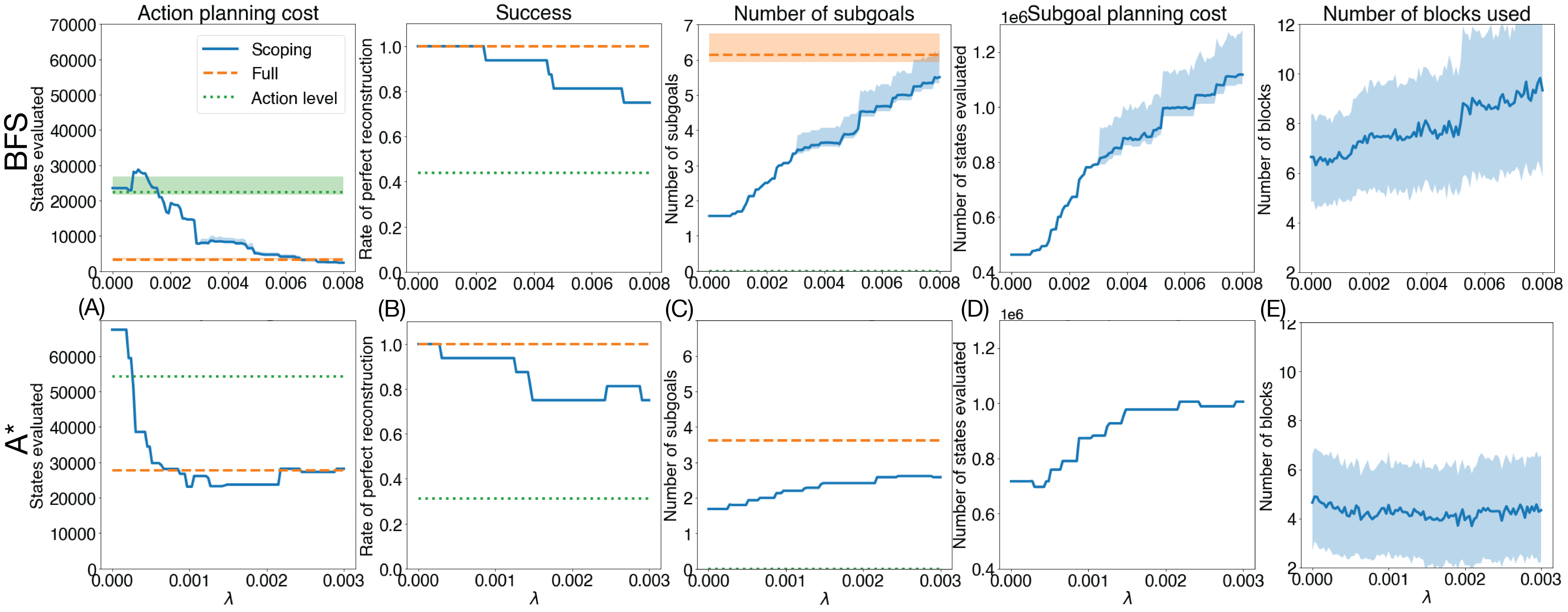}
	\end{center}
	\vspace{-6mm}
	\caption{
		Parameter $\lambda$ governs the tradeoff between choosing a subgoal that is easy to solve and one that maximizes progress. 
		(A) A higher emphasis on minimizing action planning cost as opposed to maximizing progress (increasing $\lambda$) leads to a decrease in action planning cost, 
		(B) a lower rate of perfect reconstruction,
		(C) an increase in the number of subgoals the planner actually ends up using,
		(D) as well as the total subgoal planning cost, since the scoping planners need to find the next subgoal more often.
		(E) It increases the number of blocks used for BFS lookahead, but not for A* lookahead.
		} 
	\label{lambda}
\end{figure*}

To analyze the tradeoff between avoiding costs and making progress, we apply the scoping planner under a range of values for $\lambda$---see Figure \ref{lambda}.
As expected, valuing reducing planning cost indeed results in a reduction in action planning cost 
(BFS: Pearson's r($62$) = \numns{-0.927056312707038}, $95\%$ CI: $[\numns{-0.92898624},\numns{-0.92530724}]$, $p<0.001$;
 A*: r($62$) = \numns{-0.5806041152273578}, $95\%$ CI: $[\numns{-0.58060412},\numns{-0.58060412}]$, $p<0.001$, all confidence intervals bootstrapped). 
Along with this comes a reduction in success 
(BFS: r($62$) = \numns{-0.9548686841936133},  $95\%$ CI: $[\numns{-0.95486868},\numns{-0.95486868}]$, $p<0.001$; 
 A*: r($62$) = \numns{-0.8398525744295534},  $95\%$ CI: $[\numns{-0.83985257},\numns{-0.83985257}]$, $p<0.001$): 
with lower action planning costs comes an increased chance of building oneself into a corner. 
When increasing $\lambda$, the number of subgoals also increases 
(BFS: r($62$) = \numns{0.9818182275488507}, $95\%$ CI: $[\numns{0.97631014},\numns{0.9860201}]$, $p<0.001$;
 A*: r($62$) = \numns{0.9642896556795777}, $95\%$ CI: $[\numns{0.96428966},\numns{0.96428966}]$, $p<0.001$. 
Paradoxically, trying to minimize action planning costs increases the total subgoal planning cost 
(BFS: r($62$) = \numns{0.9610956467307686}, $95\%$ CI: $[\numns{0.94980528},\numns{0.97013698}]$, $p<0.001$; 
A*: r($62$) = \numns{0.9202841132251751}, $95\%$ CI: $[\numns{0.92028411},\numns{0.92028411}]$, $p<0.001$. 
This is due to the increase in the number of subgoals, as smaller subgoals tend to be easier to plan. More subgoals means that the costly subgoal planning cost needs to be performed more often. This illustrates that cognitive resource limitations can lead to a larger total of cognitive resources used, albeit in smaller increments.

Finally, the found solutions themselves qualitatively differ across values of $\lambda$ as well as action-level search algorithms. 
Increasing $\lambda$ leads to more blocks being placed when using BFS as a action-level search algorithm, but a slight decrease with A*
(BFS: r($62$) = \numns{0.5602527483405264}, $95\%$ CI: $[\numns{0.420267},\numns{0.67768439}]$, $p<0.001$; 
A*: r($62$) = \numns{-0.05443691959186608}, $95\%$ CI: $[\numns{-0.24758946},\numns{0.13160865}]$ $p=\numns{0.143}$). 
Valuing making progress over avoiding planning cost thus leads the scoping planner with BFS lookahead to shorter solutions, whereas avoiding costs leads to more conservative, longer solutions. 


\section{Discussion}
We found that both scoping and full-subgoal planning outperformed pure action-level search, consistent with the notion that decomposing complex problems can be beneficial. 
While full-subgoal planning succeeds in finding the best task decompositions, the cost of finding this decomposition into subgoals is much higher than the scoping planner requires to achieve comparable task performance. 
In order to be a plausible proposal for how people approach these planning problems, both kinds of costs need to be taken into account. 
Here we assume that subgoal planning cost as defined is a plausible proxy for the actual cost of finding subgoals: while humans likely don't fully search for actions for all potential subgoals, the cost of subgoal planning arguably depends on the number of potential subgoals as well as their difficulty. 
Taken together, our findings suggest that visual scoping may be a promising way to reap the benefits of utilizing subgoals while minimizing the overhead of subgoal planning, thereby making efficient use of limited cognitive resources. 


Given the way that scoping exploits spatial information to manage the computational overhead involved in planning, it may also be a useful source of insight into how people solve similar spatial reasoning problems. 
Towards this end, we are developing a novel behavioral paradigm allowing people to visually define subgoals during planning, enabling us both to make inferences about cognitive resource constraints and to observe how they interleave planning and action over time. 
A promising direction for future work is the evaluation of a broader array of scoping strategies, such as those forms of scoping closer to full-subgoal planning that plan more than one subgoal into the future at a time before taking action. 
Likewise, we aim to extend visual scoping from simple two-dimensional grid worlds to richer three-dimensional environments, where effective use of cognitive resources is crucial.
In the block tower reconstruction task, the environment is deterministic. When the effect of the actions are uncertain---when the environment is stochastic or its dynamics aren't fully known to the planner---the benefit of only planning subgoals into the near future is likely going to be more relevant. A future direction is to investigate how people scope in non-deterministic environments. 
Finally, visual scoping understands the planners as embedded in space and time: the planner exploits the structure of its environment to propose subgoals, and those subgoals are proposed on the basis of actions the planner has taken before. In this sense, visual scoping connects classical algorithmic models of problem solving with the notion of cognition being always situated in and dependent on an environment \cite{KirshProblemSolvingSituated2009}.  

\section{Acknowledgments}
The authors would like to thank Will McCarthy for his development of the block tower reconstruction task, as well as members of the Cognitive Tools Lab at UC San Diego for their comments and support. 
This work was supported by NSF CAREER Award \#2047191 to J.E.F.

\vspace{2em}
\fbox{\parbox[b][][c]{8.0cm}{\centering {All code and materials available at: \\ \href{https://github.com/cogtoolslab/tools_block_construction}{\url{https://github.com/cogtoolslab/tools_block_construction}}}}}
\vspace{2em} \noindent

\bibliographystyle{apacite}

\setlength{\bibleftmargin}{.125in}
\setlength{\bibindent}{-\bibleftmargin}

\bibliography{cogsci}

\end{document}